\documentclass[11pt,a4paper]{article}
\usepackage[hyperref]{acl2020}
\usepackage{times}
\usepackage{latexsym}
\usepackage{multirow}

\usepackage{graphicx}

\usepackage{microtype}

\aclfinalcopy 

\title{Emotion Conditioned Creative Dialog Generation}

\author{Khalid Alnajjar\\
  University of Helsinki \\ Finland \\
  \texttt{khalid.alnajjar@helsinki.fi} \\\And
  Mika Hämäläinen \\
  University of Helsinki \\
  Finland \\ \\
  \texttt{mika.hamalainen@helsinki.fi } \\}

\date{}

\begin{document}
\maketitle
\begin{abstract}
We present a DialGPT based model for generating creative dialog responses that are conditioned based on one of the following emotions: \textit{anger, disgust, fear, happiness, pain, sadness} and \textit{surprise}. Our model is capable of producing a contextually apt response given an input sentence and a desired emotion label. Our model is capable of expressing the desired emotion with an accuracy of 0.6. The best performing emotions are \textit{neutral}, \textit{fear} and \textit{disgust}. When measuring the strength of the expressed emotion, we find that \textit{anger}, \textit{fear} and \textit{disgust} are expressed in the most strong fashion by the model.
\end{abstract}

\section{Introduction}

Dialog systems and different kinds of natural language interfaces are all around us. When we seek information or contact customer support, we are increasingly more often first greeted by a bot rather than a person. Bots are typically stiff and not lifelike making communication with them an awkward experience. This is because their goal oriented nature typically sets some constraints in terms of how creative a system can be.

We can perceive a gap between dialog systems that are designed to convey a certain message in a goal oriented fashion and dialog systems that generate chit-chat. Chit-chat can be generated rather freely because any topical response is valid, where as a certain degree of factual correctness is to be expected from a goal oriented system. 

In this paper, we seek to bring the two lines of dialog generation research closer together. We implement a system that can generate topical responses (in the sense of chit-chat) with a fixed emotional content. Thus the goal of the system is to convey a desired emotion in its response, no matter what the actual textual content ended up being. This means that part of the semantics of the out is fixed, while a part as to how to contextually adapt the emotional content. is still up to computational creativity.

We base our experiments on a recently published dialog dataset that contains sentiment annotations\footnote{https://zenodo.org/record/6990638}. The dataset is based on a video game called Fallout New Vegas and it has dialog where each line of the dialog is annotated as containing one of the following emotions: \textit{anger, disgust, fear, happiness, neutral, pain, sadness} or \textit{surprise}. This dataset makes for an optimal training data for the task we seek to solve.

\begin{table*}[ht]
\centering
\begin{tabular}{|l|l|l|l|l|l|l|l|l|}
\hline
          & Anger & Disgust & Fear & Happy & Neutral & Pained & Sad  & Surprised \\ \hline
Sentences & 3 335  & 932     & 1 620 & 4 029  & 8 802    & 994    & 1 055 & 1 649      \\ \hline
\end{tabular}
\caption{Number of sentences per emotion}
\label{tab:datasize}
\end{table*}

\begin{table*}[ht]
\centering
\begin{tabular}{|l|l|l|}
\hline
Prompt                              & Response                                                     & Emotion   \\ \hline
I hear you've been causing trouble. & Oh yeah? Fuck off, asshole.                                  & Anger     \\ \hline
I was hoping you'd be that stupid.  & What? Hey, guys! Help me here!                               & Surprise  \\ \hline
I've dealt with those newcomers.    & I take care of those who help with that. Here, you earned it & Happiness \\ \hline
\end{tabular}
\caption{Examples of the training data}
\label{tab:data_examples}
\end{table*}

\section{Related Work}

In terms of computational creativity and natural language generation, there are several papers out there that present work conducted on a variety of different creative language generation tasks such as poem generation \cite{hegade2021po,hamalainen2022modern}, humor generation \cite{weller2020can,alnajjar2021computer}, news generation \cite{shu2021fact,koppatz2022automatic} and story generation \cite{vicente2018statistical,concepcion2019evolving}. In this section, we will take a closer look at the work conducted on dialog generation.

\citet{xie2021empathetic} present work on generating empathetic dialog. Their model deals with the following categories of empathetic intent: \textit{questioning, agreeing, acknowledging, sympathizing, encouraging, consoling, suggesting} and \textit{wishing}. They base their model on the transformer architecture and they use RoBERTa for input encoding. In addition, they train a classifier that predicts the salient empathetic intent.

Dialog generation has also been tackled in a context-controlled and topic-aware manner \cite{ling2021context}. Their model consists of four parts: a hierarchical context encoder, a contex dependent topic representation module, a context guided topic transition module and a joint attention based response decoder.

\citet{chen2021reasoning} present their work on multi-turn dialog generation. They use a cross-hierarchical encoder that encodes a sentence for an answer selector model, after this a response generator model is used to generate the final output. The initial encoding is done by a transformer based model while the final genration is done by an LSTM model.

Dialog adaptation has been studied before in the context of video games \cite{hamalainen2019creative}. The authors use an LSTM model to paraphrase the syntax of existing dialog to introduce diversity and a word2vec model to adapt the meaning of the sentence towards a desired player attribute.

\section{Creativity and Emotion}

There are several takes on creativity in a computational setting. In this section, we cover some of these theoretical ways of understanding computational creativity. Theoretical foundation has been, for a long time, at the very core of computational creativity to combat systems that do mere generation. That is generation for the sake of outputting something by any means necessary.

A computationally creative system should exhibit skill, imagination and appreciation according to \citet{colton2008creativity}. He argues that all of these three components are a strict requirement for creativity to exist in a system. Skill refers to the system's capability of producing a creative artifact whereas appreciation means that the system should also know why its creation is good. Imagination requires the system to be capable of garnering a lot of diverse output for one input.

Creativity can also be modeled through the FACE theory \cite{colton2011computational}. This theory states creativity comes from the interplay between framing, aesthetics, concept and expression. Expressions are the creative output produced by the system. The system itself is called concept. Aesthetics is similar to appreciation in the previous theory; it means that the system should be able to appreciate the creative value of its output. Framing highlights the fact that creativity does not take place in a vacuum but is presented in a context. In our case, framing would be the entire dialog between a human user and the machine.

\citet{boden1998creativity} identifies three types of creativity; exploratory creativity, transformational creativity and combinatory creativity. In combinatory creativity, a system forms new artifacts by combining old ones in novel ways. In exploratory creativity, a system is conducing a search in a conceptual space discovering new creative artifacts. A system that can achieve transformational creativity can change its search space.

A system is considered to be autonomously creative if it can change its own standards without being explicitly told to do so \citet{jennings2010developing}. The change cannot occur at random either because a simple random change at random intervals would otherwise be enough to satisfy the criterion.

Emotion is considered as a higher level cognitive phenomenon than a feeling (see \citealt{shouse2005feeling}). Feelings are seen to be universally felt in a similar fashion as a response to some external or internal stimulus. Emotions, on the other hand, are culturally and socially represented and their existence in all cultures in a similar way is not a given thing (see \citealt{lim2016cultural}). 

One way of seeing emotions is that they rely on affect (see \citealt{russell2003core}), which is a state our mind is continuously in. An affect can move across two axes: positive-negative and arousal-relaxed. The affect we feel is contextually resolved to a higher level emotion based on the context we find ourselves in. For instance, a high arousal and negativity could be interpreted either as anger or disgust among others.

\citet{ekman1992there} has identified six basic emotions: \textit{anger, disgust, fear, happiness, sadness} and \textit{surprise}. These emotions are also present in our dataset, which makes this theory optimal to build upon. The basic emotions are considered to be universal across cultures based on studies conducted on facial expressions.

\begin{table*}[ht]
\centering
\begin{tabular}{|l|l|l|}
\hline
Prompt                                                                                                                                          & Emotion  & Output                                                                                            \\ \hline
Doesn't look like the talkative type.                                                                                                           & Anger    & What the fuck are you looking at?                                                                 \\ \hline
Wait a minute! I ain't done nothin' to you!                                                                                                     & Fear     & I don't want any trouble, okay?                                                                   \\ \hline
Asked you a question. How many?                                                                                                                 & Sad      & \begin{tabular}[c]{@{}l@{}}I'm sorry, but I don't have the time\\ to talk right now.\end{tabular} \\ \hline
\begin{tabular}[c]{@{}l@{}}Need coin, fast? Those blasted nonhumans at the\\  Vivaldi Bank say your credit's no good? Come see me!\end{tabular} & Surprise & I've never heard of that before.                                                                  \\ \hline
\end{tabular}
\caption{Examples of the results}
\label{tab:results_examples}
\end{table*}

\section{Data}

Our dataset consists of dialog in English where each line is annotated with an emotion label. Table \ref{tab:datasize} shows the data size and how many sentences were there in the dataset for each emotion. The dialog is from a video game called Fallout New Vegas\footnote{https://fallout.bethesda.net/en/games/fallout-new-vegas} by Obsidian Entertainment. The game is set in a post-apocalyptic world inhabitet by creatures that were born as a result of nuclear radiation such as mutants, ghouls and feral ghouls in addition to humans.

The game dialog consists mostly of player prompts that result in a response by an NPC (non-player character). In Table \ref{tab:data_examples} we can see some examples of the dialog. In our case, we train our model by using the prompt and the emotion label as an input to predict the response.

Fallout New Vegas is a relatively large video game because it is an open world RPG (role-playing game) where the player can roam freely from one place to another. As a result of this, the game has a variety of different scripted characters which means that there is no bias in terms of having the same characters speaking with each other all the time. There is, however, a bias in the topic of conversations given that they take place in a fictional world. Many of the dialogs deal with fictional places, characters, items and so on.

\section{Dialog Generation}

In this section, we outline our approach to emotionally conditioned dialog generation. We conduct our experiment using Transformers \cite{wolf-etal-2020-transformers} and Datasets \cite{lhoest-etal-2021-datasets} Python libraries. We base our model on a pretrained model called DialoGPT medium\footnote{https://huggingface.co/microsoft/DialoGPT-medium}.

DialoGPT \cite{zhang2020dialogpt} is based on the GPT-2 \cite{radford2019language} architecture, which in turn is based on the generic transformer language model \cite{vaswani2017attention}. The transformer model leverages a stack of masked multi-head self-attention layers to train on large datasets. DialoGPT employs a maximum mutual information scoring function \cite{li2015diversity,zhang2018generating} during the training phase optimizing the reward with a policy gradient \cite{williams1992simple} with a sample averaged baseline \cite{zhang2018generating}.

We changed the format of the data to be as ``EMOTION$_{1}$: SENTENCE$_{1}$. EMOTION$_{2}$: SENTENCE$_{2}$ [EOS]``, where EMOTION$_{n}$ indicates the emotion of the $n$th sentence in the conversation. This way, the model is exposed to emotional knowledge regarding the preceding locution of the response, as well. We use 90\% of the data for training and 10\% for validation. We train the model for 5 epochs.

\section{Results and Evaluation}

In order to evaluate our model, we sample dialog from an unrelated open-world RPG called The Witcher 3: Wild Hunt\footnote{https://www.thewitcher.com/en/witcher3} by CD Projekt. The dialog is extracted using a w3strings decoder tool\footnote{https://www.nexusmods.com/witcher3/mods/1055/}. By using dialog from a different video game, we can see whether our model works in a different domain. The Witcher 3 is set to a medieval fantasy world inhabited by magical creatures. Some of the generated output can be seen in Table \ref{tab:results_examples}. 

We sample randomly 15 sentences for each emotion, and produce emotion conditioned replies. In the input, ``EMOTION$_{1}$:`` was not present as the dialogs in The Witcher did not include emotion labels. These are then evaluated by crowd-workers on a crowd-sourcing platform called Appen\footnote{https://appen.com/}. For every output we evaluate, we show the input sentence and the output to the user and ask them if the output expresses the desired emotion as a simple yes/no question. In case, the user selected "yes", we also asked how strongly the line expressed the desired emotion on a 5-point scale (the stronger, the higher), the scale is 0-based from 0 to 4. Each generated output was evaluated by 5 different judges. 

\begin{figure}[h]

\centering
\includegraphics[width=0.5\textwidth]{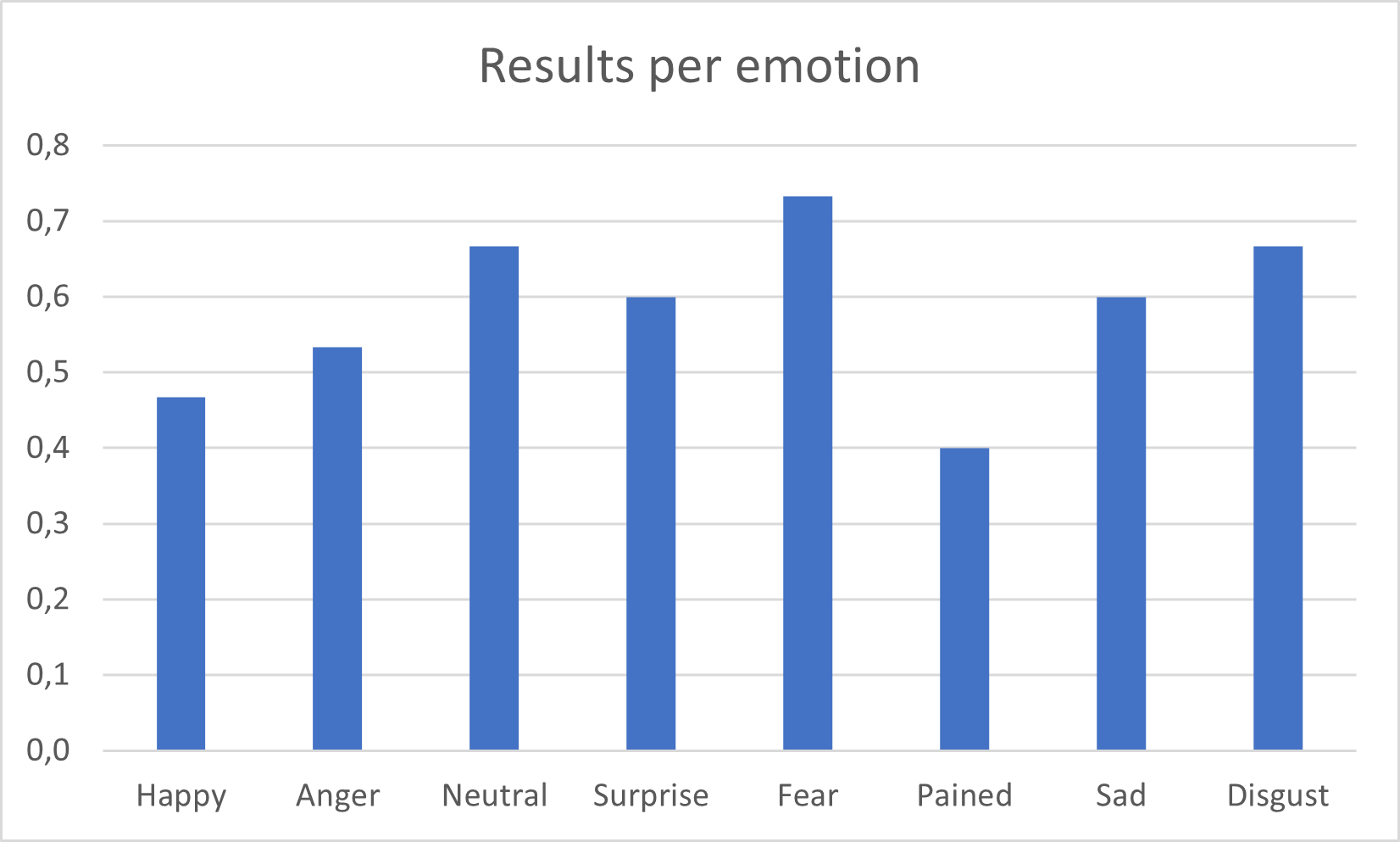}
\caption{Percentage of "yes" answers per emotion}
\label{fig:yes_answers}
\end{figure}

All in all, the judges agreement with the emotion label was \textbf{0.6}. In Figure \ref{fig:yes_answers}, we can see the judges' agreement per emotion label. The worst performing emotion is \textit{pained} and the best performing ones are \textit{neutral}, \textit{fear} and \textit{disgust}. \textit{Sad}, \textit{surprise} and \textit{anger} are all rated expressing the emotion over half of the time. These results are based on the aggregated results provided by Appen.

\begin{figure}[h]

\centering
\includegraphics[width=0.5\textwidth]{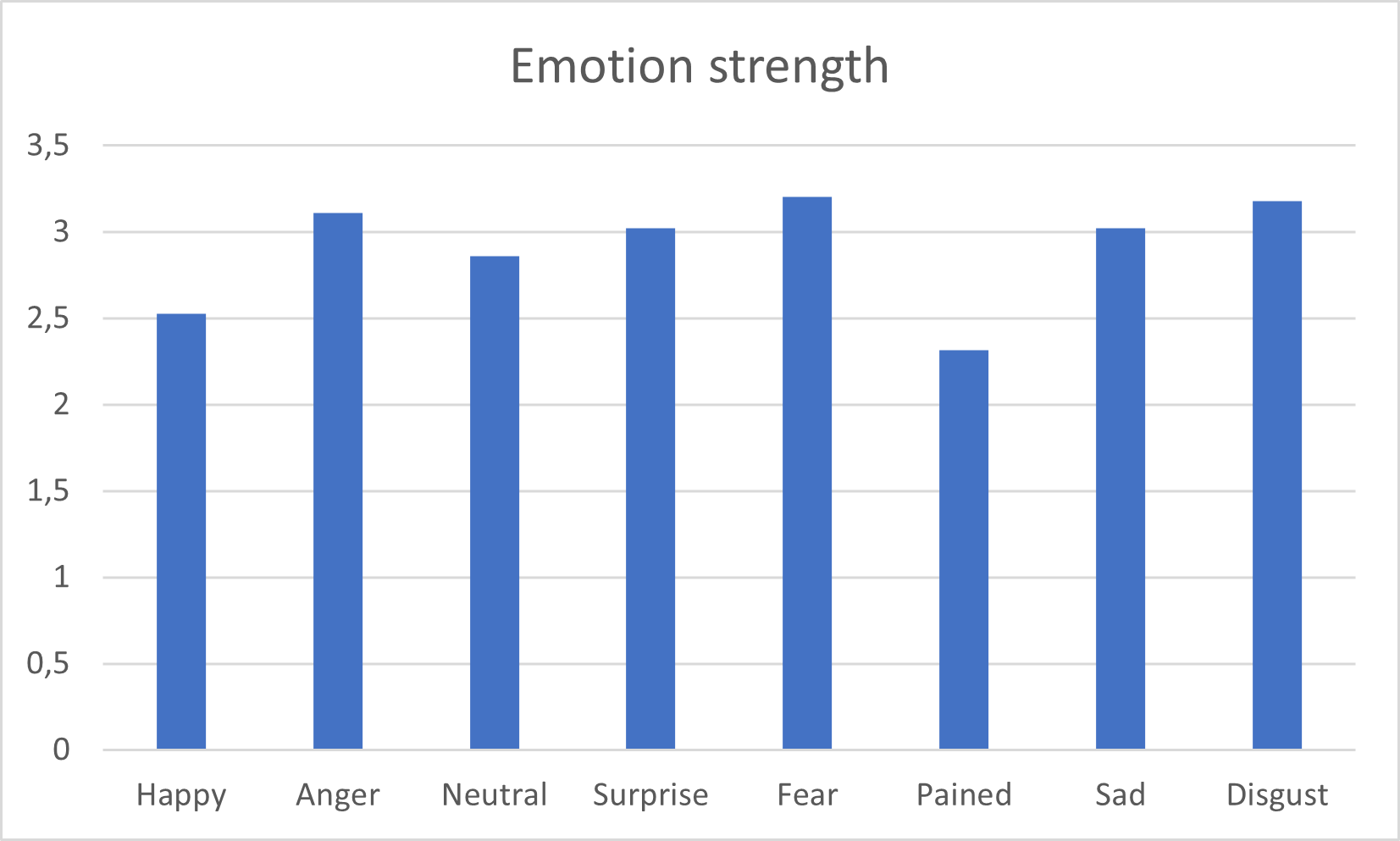}
\caption{Strength of each emotion}
\label{fig:strength}
\end{figure}

Figure \ref{fig:strength} shows how strong the judges rated each emotion to be. 0 indicates not strong at all and 4 very strong. As we can see, all emotions achieve above mid-level performance. \textit{Anger, fear} and \textit{disgust} being the most strongly expressed emotions. All in all, based on the results, it seems that the model is capable of producing emotional responses in a suitably strong manner.

\section{Conclusions}

Emotion conditioned dialog generation remains a challenge. In this paper, we have shown the viability of using a DialoGPT model for this task. The results are promising especially given that the evaluation was run on out-of-the-domain data.

The video game that was used in this paper also comes with audio files for each line of dialog. This is an interesting direction for future research because we could generate emotion conditioned dialog with audio as well. The same sentence can be said with multiple different tones and intonations to express different emotions. Building a system that can express emotion in generated speech as well would have great application potentials in speech oriented dialog systems.

\section*{Acknowledgments}
This work was partially financed by the Society of Swedish Literature in Finland with funding from Enhancing Conversational AI with Computational Creativity, and by the Ella and Georg Ehrnrooth Foundation for Modelling Conversational Artificial Intelligence with Intent and Creativity.

\bibliography{anthology,acl2020}
\bibliographystyle{acl_natbib}

\end{document}